\documentclass[a4paper, UKenglish, cleveref, autoref, thm-restate]{lipics-v2021}
\pdfoutput=1
\hideLIPIcs

\graphicspath{{figs/}}

\title{Accelerated Fourier SAT: Fully Realising a GPU-based Symmetric Pseudo-Boolean SAT Solver}

\titlerunning{Accelerated Fourier SAT: GPU-based Pseudo-Boolean Solving}

\author{Cody J. Christopher}{School of Computing, Australian National University, Canberra, Australia.}{cody.christopher@anu.edu.au}{https://orcid.org/0000-0001-8444-2292}{}

\author{Charles Gretton}{School of Computing, Australian National University, Canberra, Australia.}{charles.gretton@anu.edu.au}{https://orcid.org/0000-0001-9803-0168}{}

\authorrunning{C.\,J. Christopher and C. Gretton}

\Copyright{Cody J Christopher and Charles Gretton}



\supplement{\afs{} implementation with \texttt{Apache-2.0}/\texttt{GPL-2.0-or-later} licenses:}
\supplementdetails[linktext={},
                   cite={},
                   subcategory={Source Code}, 
                   swhid={}
                   ]{Software}{https://github.com/cjchristopher/accelerated-fourier-sat}

\nolinenumbers

\EventEditors{}
\EventNoEds{0}
\EventLongTitle{}
\EventShortTitle{}
\EventAcronym{}
\EventYear{}
\EventDate{}
\EventLocation{}
\EventLogo{}
\SeriesVolume{}
\ArticleNo{}

\hypersetup{
colorlinks=true, 
breaklinks=true, 
bookmarksnumbered,
pdftitle={}, % PDF title
pdfauthor={\textcopyright}, % PDF Author
pdfsubject={}, % PDF Subject
pdfkeywords={}, % PDF Keywords
pdfcreator={pdfLaTeX}, % PDF Creator
pdfproducer={LaTeX with hyperref and ClassicThesis} % PDF producer
}

\usepackage[T1]{fontenc} % Use 8-bit encoding that has 256 glyphs

\usepackage[utf8]{inputenc} % Required for including letters with accents
\usepackage{fix-cm}

\usepackage{graphicx} % Required for including images

\usepackage{lipsum} % Used for inserting dummy 'Lorem ipsum' text into the template

\usepackage{subcaption}

\usepackage{amsmath,amssymb,amsthm,latexsym} % For including math equations, theorems, symbols, etc

\usepackage{bm} % bold math.

\usepackage{semantic} %math ligatures

\usepackage[shortcuts]{extdash} % manual hyphenation and explicit dash variations (em \---, en \--)

\usepackage{xspace} % smart spacing mostly for macros}

\usepackage{cellspace}
\usepackage{makecell}
\usepackage[dvipsnames]{xcolor}
\usepackage{fancyvrb}
\usepackage{tikz}
\usepackage{tikzscale}
\usepackage{pgfplots}
\usepackage{graphicx}
\usepackage[ruled,vlined]{algorithm2e}
\usepackage{hyperref}
\usepackage{tcolorbox}
\tcbuselibrary{listings,breakable}
\usepackage{mathtools}
\usepackage[inkscapelatex=false]{svg}

\usepackage{listings} % arXiv-safe code blocks
\usepackage{upquote} % straight quotes in verbatim/listings

\providecommand{\abs}[1]{\ensuremath{\left\lvert#1\right\rvert}}

\providecommand{\max}[1]{\ensuremath{\max\left(#1\right)}}
\providecommand{\min}[1]{\ensuremath{\min\left(#1\right)}}

\providecommand{\mset}[1]{\ensuremath{\left\{#1\right\}}}

\newcommand{\given}[0]{\text{ . }}

\mathlig{|}{\mid}
\mathlig{==}{\equiv}
\mathlig{/==}{\not\equiv}
\mathlig{~>}{\leadsto}
\mathlig{~=}{\approx}
\mathlig{~}{\lnot}
\mathlig{|->}{\mapsto}
\mathlig{||-}{\Vdash}
\mathlig{|/=}{\nvDash}
\mathlig{|=}{\vDash}
\mathlig{|/-}{\nvdash}
\mathlig{?}{\mathbin{?}}
\mathlig{||}{\parallel}
\mathlig{<=>}{\Leftrightarrow}
\mathlig{<==>}{\Longleftrightarrow}
\newcommand\etc{etc.\xspace}
\newcommand{\ie}{i.e.\xspace}
\newcommand{\eg}{e.g.\xspace}

\newcommand\extrafootertext[1]{%
    \bgroup
    \renewcommand\thefootnote{\fnsymbol{footnote}}%
    \renewcommand\thempfootnote{\fnsymbol{mpfootnote}}%
    \footnotetext[0]{#1}%
    \egroup
}

\definecolor{BashBG}{HTML}{0B0F14}
\definecolor{BashFG}{HTML}{E6E6E6}
\definecolor{BashAccent}{HTML}{7FDBCA}

\lstdefinestyle{bashstyle}{
  language=bash,
  basicstyle=\ttfamily\footnotesize,
  breaklines=true,
  columns=fullflexible,
  keepspaces=true,
  showstringspaces=false,
  upquote=true,
  keywordstyle=\color{NavyBlue}\bfseries,
  commentstyle=\color{OliveGreen},
  stringstyle=\color{BrickRed},
  backgroundcolor=\color{white}
}

\newtcblisting{bashblock}{
  listing engine=listings,
  listing only,
  breakable,
  colback=white,
  colframe=black,
  boxrule=0.5pt,
  arc=0pt,
  outer arc=0pt,
  left=0mm,
  right=0mm,
  top=0mm,
  bottom=0mm,
  boxsep=2mm,
  listing options={style=bashstyle}
}

\newtcolorbox{terminalbox}[1][]{
  colback=BashBG,coltext=BashFG,
  colframe=BashAccent, fonttitle=\bfseries,
  listing only, breakable, enhanced,
  left=1mm, right=1mm, top=1mm, bottom=1mm,
  listing options={language=bash,breaklines=true,basicstyle=\ttfamily\footnotesize},
  title=#1
}

\usetikzlibrary{arrows.meta}
\usetikzlibrary{arrows}
\usepgfplotslibrary{fillbetween}

\NewDocumentCommand{\FE}{o}{%
  \IfNoValueTF{#1}{%
    \texttt{FE}%
  }{%
    \ensuremath{\texttt{FE}_{#1}}
  }%
}

\newcommand{\afs}{\texttt{AFSAT}}
\pgfplotsset{compat=1.18}

\begin{document}
\date{\today}
\maketitle

%----------------------------------------------------------------------------------------
%  ABSTRACT
%----------------------------------------------------------------------------------------
\begin{abstract}
We present Accelerated Fourier SAT (\afs{}), a GPU-accelerated solver for pseudo-Boolean satisfiability based on continuous local search (CLS).
\afs{} realises the proof-of-concept approach, \texttt{FastFourierSAT}, into a fully-engineered solver supporting any heterogeneous mixture of symmetric constraint types and lengths within a single problem instance.
Using the \textsc{JAX} compiler, \afs{} leverages pure function composition, automatic vectorisation, automatic differentiation, and just-in-time (JIT) compilation to perform massively parallel CLS across batches of candidate assignments.
We demonstrate substantially improved numerical stability, runtime performance, and memory efficiency over the proof-of-concept. We achieve this by way of identifying and addressing various limitations that arise from memory latency and floating-point representation, as well as leveraging automatic parallelisation and compact representations. The inherent representational and stability limitations of floating point are partially addressed by a tailored discrete Fourier transform implementation. We achieve near-linear throughput when scaling to multiple accelerators via \textsc{JAX} array sharding.
\end{abstract}

%----------------------------------------------------------------------------------------
%  INTRODUCTION
%----------------------------------------------------------------------------------------
\section{Introduction}

Continuous local search (CLS) offers a compelling search paradigm for solving satisfiability (SAT) problems that are expressed using symmetric pseudo-Boolean (PB) constraints. The approach relaxes Boolean problem variables to real-valued variables by way of the Walsh-Fourier transform from Boolean function analysis~\cite{odonnell14}. Revisiting this approach, SAT is reformulated as a bounded continuous optimisation problem amenable to gradient-based search methods. This approach was formalised for the \texttt{FourierSAT} proof-of-concept~\cite{kyrillidis2021solving} and subsequently extended for parallelised GPU computation in \texttt{FastFourierSAT}~\cite{cen2025massively}, which demonstrated that the Walsh-Fourier expansion can be evaluated efficiently using a vectorised Discrete Fourier Transform (DFT) in $O^{*}(\log k)$ time where $k$ is the number of variables (typically also literals) in a constraint (clause), and $O^{*}(\cdot)\stackrel{def}{=} O_{p->\infty}(T_p(\cdot))$ is idealised parallel execution time with infinite resources.

We present Accelerated Fourier SAT (\afs{}), a ground-up re-engineered and extended implementation of this CLS approach. \afs{} provides the following improvements and novel contributions as a tool:
\begin{itemize}
  \item \textbf{Heterogeneous constraint support.} \afs{} is the first CLS implementation to support problems with an arbitrary mix of common PB constraint types
  of varying lengths within a single problem instance.
  \item \textbf{Improved performance and efficiency.} We demonstrate better execution times (up to parity in the worst case) with less overhead, substantially reduced GPU memory consumption, and higher peak throughput (evaluations/searches per unit time) compared to \texttt{FastFourierSAT}.
  \item \textbf{Multi-GPU scaling.} We demonstrate near-linear scaling across multiple GPUs using distributed array sharding in \textsc{JAX}~\cite{deepmind2020jax}, in a single-program-multiple-data (SPMD), or compute-follows-data paradigm.
  \item \textbf{Numerical stability and precision improvements.} We identify and address sources of floating-point errors and representational deviations through a tailored DFT matrix construction with deferred division. This establishes a practical maximum constraint length of approximately 50 variables.
  \item \textbf{Partial assignment integration.} \afs{} accepts partial variable assignments as input, opening up avenues to use it as a sub-solver within decomposition-based architectures such as \textsc{Dagster}~\cite{10.1007/978-3-031-20862-1_6}, or other portfolio approaches.
\end{itemize}

%----------------------------------------------------------------------------------------
%  BACKGROUND
%----------------------------------------------------------------------------------------

\section{Background}
\label{sec:background}

\subsection{Continuous Relaxation via Walsh-Fourier Expansion}
CLS operates on a continuous relaxation of a SAT problem. For a Boolean formula $\phi$ over $n$ variables, we seek a relaxation as a polynomial on the Boolean hypercube $\mathcal{Q}^n$ that preserves satisfiability (\ie a solution to the relaxed formulation provides a solution for the discrete problem). The (Walsh-)Fourier expansion ($\FE[]$)~\cite{odonnell14} of Boolean functions provides this relaxation, mapping $\mset{\texttt{True},\texttt{False}}$ to $\mset{-1,1}$:
\begin{equation}
  \label{eqn:wft}
  \FE[\phi](\mathbf{X}) \stackrel{\text{def}}{=} \sum_{S\in 2^{\mathbf{X}}}\hat{f}(S)\prod_{x_i\in S} x_i
\end{equation}
where $\hat{f}(S)$ are called the Fourier coefficients. For \emph{symmetric} constraints---those whose truth value depends only on the \emph{simple count} of true literals---closed-form solutions for the coefficients exist that are computable in polynomial time at worst~\cite{kyrillidis2021solving}. Symmetric constraints necessarily have idempotent variable weights.

\subsubsection{Satisfaction as Optimisation}
Given a formula $\phi$ in $n$ variables, decomposed into $m$ symmetric constraints $C_1,\ldots,C_m$, the satisfiability problem is expressed as the bounded minimisation:
\begin{equation}
  \min_{\mathbf{X}} \sum_{k=1}^{m}\FE[C_k](\mathbf{X}) \text{ subject to } \mathbf{X} \in \mathcal{Q}^n
\end{equation}
where an assignment $\mathbf{X} \in [-1,1]^n$ \emph{satisfies} $\phi$ if $\sum_k \FE[C_k](\mathbf{X}) = -m$. This formulation is differentiable, non-convex, bounded, and saddle-dense. This motivates the use of projected and/or bounded gradient search methods as the first choice of search algorithms. Observing that $\FE[]$ is differentiable and fast parallelised automatic differentiation is available, we would also consider second- or higher-order methods, should they exist.

\subsubsection{DFT-Based Vectorised Evaluation}
A key insight from Cen et al.~\cite{cen2025massively} is that the evaluation of the Fourier expansion can be performed using a DFT. We observe that the terms in $\FE[\phi]$ sharing a coefficient $\hat{f}(S)$ are all the size $\abs{S}$ combinations of the variables of the clause. These particular sums are well-studied polynomials known as the \emph{elementary symmetric polynomials} (ESPs).
Since the sequence of all length $k$ ESPs up to $n$ variables, $\bm{e}^n_k$, can be computed with ($0$-padded) linear convolutions, we can compute the evaluation of a Fourier expanded formula with vectorised operations in the frequency domain by making use of the linear convolution theorem:
\begin{align*}
    \hat{f}_\phi &\equiv [\hat{f}_{\phi}(\emptyset), \hat{f}_{\phi}(1),\ldots,\hat{f}_{\phi}(n)]\\
    \bm{e}^{n} &\equiv\left [e_0^{n}, e_1^{n}, \ldots, e_{n}^{n}\right ] = \left([x_1, 1, 0,\ldots] * [x_2, 1, 0,\ldots] * \ldots * [x_{n},1, 0,\ldots]\right)\\
    &= W^{*}W\left([x_1, 1, 0, \ldots] * [x_2, 1, 0, \ldots] * \ldots * [x_n, 1, 0, \ldots]\right)\\
    &= W^{*}\left(W\left([x_1, 1, 0, \ldots]\right) \cdot W\left([x_2, 1, 0, \ldots]\right)\cdot \ldots \cdot W\left([x_n, 1, 0, \ldots]\right)\right)\\
    &= W^{*}\left([1+x_1,\omega+x_1,\ldots,\omega^n+x_1] \cdot \ldots \cdot [1+x_n,\ldots,\omega^n+x_n]\right)\\
    &= W^{*}\left(\left [1, \omega, \omega^2,\ldots,\omega^n \right ]^{T} +\; \left [x_1, x_2, \ldots, x_n \right ] \right) & \mathllap{\rhd\text{\small(Outer addition)}}\\
    &= W^{*}\left(\left[\prod_{i=1}^{n}{(1+x_i)},\prod_{i=1}^{n}{(\omega+x_i)},\ldots,\prod_{i=1}^{n}{(\omega^n+x_i)}\right]\right)\\
    &= W^{*}\left(\left[\prod_{i=1}^{n}{(\omega^j+x_i)}\right]_{j=0}^{n}\right)\\
    \FE[\phi] &\equiv \sum{\left(\hat{f}_\phi\cdot\bm{e}^{n}\right)}\\
    &= \sum{\left(\hat{f}_\phi W^{*}\cdot
    \left(\left[\prod_{i=1}^{n}{(\omega^j+x_i)}\right]_{j=0}^{n}\right)
    \right)}
\end{align*}
Where $*$ is the convolution operator on sequences, $\cdot$ is the Hadamard product (element-wise multiplication), $W$ is the DFT matrix in $n+1$ dimensions ($W^{*}$ the conjugate transpose), and $\omega$ the primitive $(n+1)^{\text{th}}$ root of unity. By taking the DFT of the ESP sequence convolution, the computation can be expressed as element-wise products and sums.

%----------------------------------------------------------------------------------------
%  SYSTEM DESIGN
%----------------------------------------------------------------------------------------

\section{System Design and Implementation}

\subsection{Architecture}

\afs{} is implemented in Python using the \textsc{JAX}~\cite{deepmind2020jax} and XLA compilers. The architecture comprises:
\begin{enumerate}
  \item \textbf{Problem ingestion}: The specification and parsing of PB-encoded problem instances specified in standard DIMACS or our hybrid variant (Appendix~\ref{sec:input_format}). 
  \item \textbf{Fourier coefficient computation}: Closed-form computation of coefficients for all supported symmetric constraint types (\S\ref{sec:constraints}).
  \item \textbf{DFT precision}: Tailored calculation of roots of unity for conjugate symmetry (\S\ref{subsec:dft}).
  \item \textbf{Compiled solver kernel}: JIT-compiled search algorithm with automatic differentiation, vectorised over batches of candidate assignments.
  \item \textbf{Multi-GPU distribution}: Sharding of batched assignments across available GPUs.
\end{enumerate}

The solver kernel is compiled once and executed repeatedly with different random initialisations. Compilation produces an optimised XLA HLO program targeting the available accelerator architecture. \afs{} exposes many additional parameters and heuristic options which can be adjusted to target specific problem types.

\subsection{Supported Constraint Types}
\label{sec:constraints}
Unlike prior CLS implementations which are restricted to either a single constraint type, or a particular fixed set of constraint types per problem, \afs{} supports heterogeneous problems containing any combination of the following symmetric pseudo-Boolean constraint types:

\begin{table}[htb]
  \centering
    \begin{tabular}{|Sl|Sl|Sl|}
      \hline
      \textbf{Constraint Type} & \textbf{PB Form} & \textbf{Coefficient Cost} \\
      \hline
      Disjunction (\texttt{OR}) & $\sum_nx_n \ge 1$ & $O(1)$ \\
      \hline
      At most one (\texttt{AMO}) & $\sum_nx_n \le 1$ & $O(n)$ \\
      \hline
      Exactly one (\texttt{EO}) & $\sum_nx_n = 1$ & $O(n)$ \\
      \hline
      Exactly $k$ (\texttt{EK}) & $\sum_nx_n = k$ & $O(n\log^{2}{n})$ \\
      \hline
      Not all equal (\texttt{NAE}) & $\bigwedge
      \begin{cases}\textstyle\sum_nx_n < n\\\textstyle\sum_nx_n > 0
      \end{cases}$ & $O(1)$ \\
      \hline
      Exclusive Or (\texttt{XOR}) & $\sum_nx_n \equiv 1 \pmod{2}$ & $O(n)$ \\
      \hline
      Cardinality-$k$ (\texttt{CARD}) & $\sum_nx_n \ge k$ & $O(n\log^{2}{n})$ \\
      \hline
  \end{tabular}
  \caption{Supported symmetric PB constraint types and asymptotic cost of computing their Walsh-Fourier expansion coefficients ($\hat{f}(S)$).}
  \label{tab:constraints}
\end{table}

This heterogeneous support enables \afs{} to handle problems with native PB formulations directly, avoiding the representational blowup of CNF translation.

\subsection{Gradient Descent and other search Algorithms}
We follow \texttt{FastFourierSAT} and select Projected Gradient Descent (PGD, Algorithm~\ref{algo:pgd}) as a fast baseline search algorithm for \afs{}. We also provide various other algorithms that possess compatible implementations. For any selected search algorithm, we take a vector-map across a batch of $B$ candidate assignments. This is realised by \textsc{JAX} in GPU warps, whereby every candidate assignment has dedicated memory and a streaming processor, and the chosen algorithm executes in lockstep across every candidate in the batch. For the choice of PGD, each descent operates numerically independently: a line search determines the step size, a gradient step is taken, and the result is projected back onto the bounded subspace $\mathcal{Q}^n$ if we happen to step out. Termination occurs when either a convergence criterion is met (e.g. new location is within $\varepsilon$ of the previous location) or given step-limit is reached. The convergence criteria is checked after every step, and some candidates in the batch may converge sooner than others. As parallelism is achieved via warps, the converged candidates execute the equivalent of no-ops until the entire batch finishes.
\begin{algorithm}[htb]
\SetAlgoLined
\KwIn{Fourier expanded formula $\FE[\phi]$, initial valuation $\textbf{X}_{(0)}$, maximum iterations $d$, convergence threshold $\delta$, maximum step-size $s$, bounds $\mathcal{Q}$}
\KwOut{Finishing assignment $\bar{\textbf{X}}$, evaluation $\FE[\phi](\bar{\textbf{X}})$, unsatisfied clause count $\#\left(\sum_{m}1 \given \FE[g_m](\bar{\textbf{X}})>0\right)$, iterations $t$}
 \For{$t\gets 1$ \KwTo $d$}{
  $\bm{\eta} <- \text{lineSearch}\left(\FE[\phi], \nabla(\FE[\phi]), s\right)$\tcp*{Determine step length}
  $\textbf{X}_{(t)} <- \textbf{X}_{(t-1)} - \bm{\eta}\cdot\nabla\left(\FE[\phi](\textbf{X}_{(t-1)})\right)$ \tcp*{Take descent step}
  $\textbf{X}_{(t)} <- \text{projectToBounds}(\textbf{X}_{(t)}, \mathcal{Q})$ \tcp*{Return to bounds}
  $is\_sat <- \texttt{True}$ \textbf{if} $\text{unsat\_count}(\phi, \textbf{X}_{(t)})=0$ \textbf{else} \texttt{False}\;
  \If(\tcp*[f]{Converged or SAT}){$\eta < \delta \text{ \textbf{or} } is\_sat$}{
   \KwRet $\textbf{X}_{(t)}, \FE[\phi](\textbf{X}_{(t)}), \text{unsat\_count}(\phi, \textbf{X}_{(t)}), t$\;
  }
 }
 \KwRet $\textbf{X}_{(t)}, \FE[\phi](\textbf{X}_{(t)}), \text{unsat\_count}(\phi, \textbf{X}_{(t)}), d$\;
 \caption{Projected Gradient Descent}\label{algo:pgd}
\end{algorithm}

\subsection{Partial Assignment Support}

\afs{} accepts a partial variable assignment as input, fixing specified variables across all batched starting valuations while randomising the remainder. \textsc{JAX} supports gradient masking (zeroing), which forces zero gradient on the relevant variables during the automatic differentiation pass resulting in a zero step length. This feature enables integration into decomposition frameworks where a partial assignment from systematic search is completed by CLS, where some orchestrator may track candidate assignments for solvers in a portfolio to improve or advise upon. \afs{} evenly divides up a given batch of candidate assignment amongst multiple partial assignment if they are provided. During clause processing, \afs{} will also detect trivial unit literals and either fold them into all provided partial assignments, or convert to a universal partial assignment.

\subsection{Floating-Point Limits on Constraint Length}
\label{subsec:fp}
The DFT-based evaluation for a constraint of length $n$ involves products $\prod_{i=1}^{n}(\omega^j + x_i)$ which, for roots $\omega^j \approx 1$ and literal values $x_i \approx 1$, can reach magnitudes of $2^n$, while for $\omega^j \approx -1$ these products approach values on the order of $\varepsilon^n$ where $\abs{\varepsilon}\to 0$. For IEEE-754 64-bit arithmetic with machine epsilon $\epsilon = 2^{-52}$, the dynamic range of these terms exceeds representable precision when $n \gtrapprox 50$, causing catastrophic cancellation in the inverse DFT, producing incorrect and impermissible evaluations $|\FE[\phi]| \gg 1$.

We empirically observe degenerate solver behaviour---exploding and vanishing gradients---for constraints of length $n \geq 48$ for which many of variables trend toward the same truth value (e.g. all but one variable assignment in an \texttt{AMO} constraint of length 50 will tend to false (1)). This establishes a practical ceiling for CLS on current GPU architectures that lack practical support for extended-precision arithmetic.

\subsubsection{Tailored DFT Construction}
\label{subsec:dft}
Standard DFT implementations (such as those from common scientific processing packages e.g., from \textsc{SciPy}~\cite{2020SciPy-NMeth}) introduce cumulative errors through repeated exponentiation of roots of unity, breaking the conjugate symmetry required for precise cancellation in the inverse DFT. Since \afs{}'s algebraic foundation is sensitive to these inaccuracies, we implement or own tailored procedures:
\begin{itemize}
  \item Forward and inverse DFT matrices are constructed to guarantee exact conjugate symmetry between paired terms.
  \item Combinatorial terms for Fourier coefficients are computed using arbitrary-precision integer arithmetic, with division and conversion to float deferred to the latest possible stage.
  \item Where algebraic symmetries exist, terms are explicitly mirrored rather than independently computed.
\end{itemize}

%----------------------------------------------------------------------------------------
%  PERFORMANCE EVALUATION
%----------------------------------------------------------------------------------------

\section{Performance Evaluation}
\label{sec:eval}
All experiments were conducted on the \textsc{Gadi} supercomputer (NCI, Australia), using nodes from the Volta GPU partition. Each node provides four NVIDIA Tesla V100 GPUs (32GB HBM2 each, $\approx$900GB/s bandwidth), Intel Xeon Cascade Lake CPUs (48 cores), and $\approx$192GB system RAM.

\subsection{Comparison with \texttt{FastFourierSAT}}

\begin{figure}[htb]
  \centering
  \begin{subfigure}{0.495\textwidth}
    \includegraphics[width=0.98\linewidth]{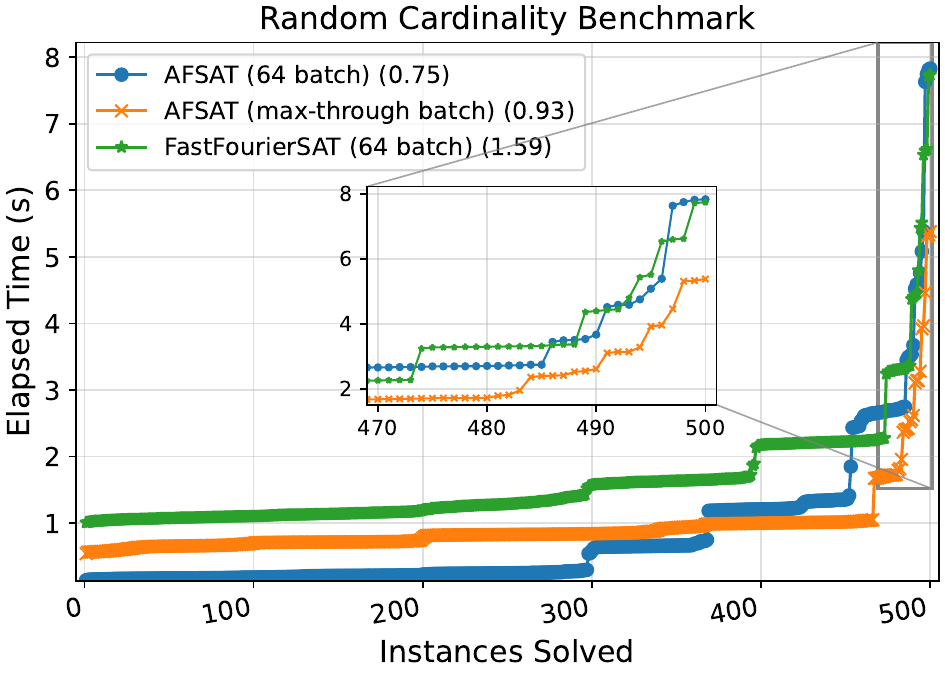}
  \end{subfigure}
  \begin{subfigure}{0.495\textwidth}
    \includegraphics[width=0.98\linewidth]{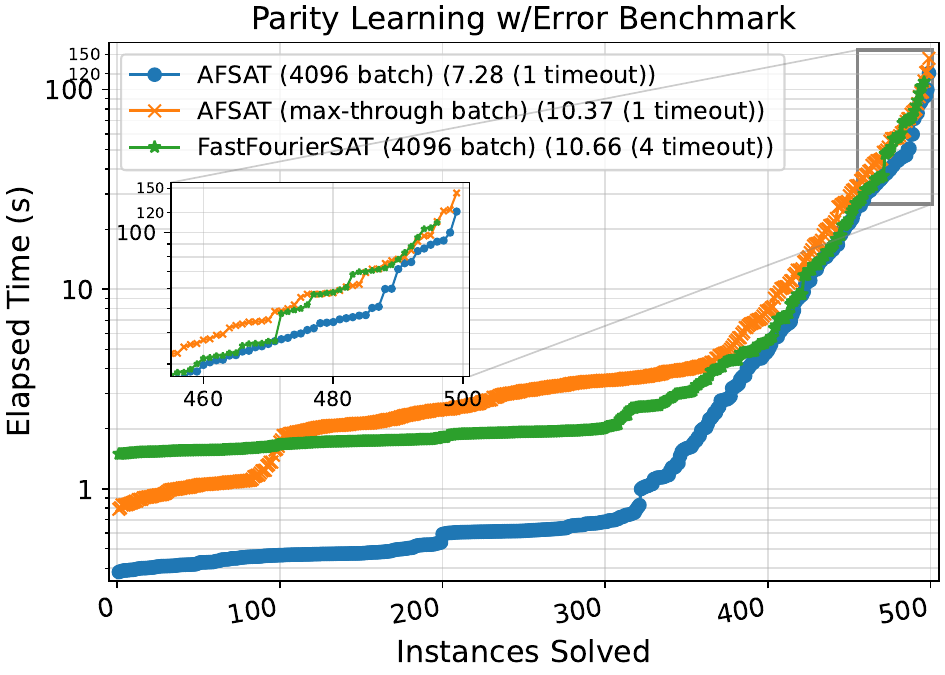}
  \end{subfigure}
  \caption{Replicated benchmarks for random cardinality \emph{(Subfig. a)} and parity learning (xor) with errors \emph{(Subfig. b)} as per \texttt{FastFourierSAT}~\cite{cen2025massively}, with PAR-2 scores and timeouts indicated. We run \afs{} configured as close to possible to \texttt{FastFourierSAT}, replicating the batch size choice as indicated in the original benchmark. We also run \afs{} in \emph{max-through} mode, where the batch size is selected to target maximum efficiency (see \S\ref{sec:gpumem}).}
  \label{fig:ffsatcard}
\end{figure}

We replicate the cardinality constraint and parity-learning benchmarks from Cen et al.~\cite{cen2025massively}. Figure~\ref{fig:ffsatcard} compares cumulative solution times. We apply certain \textsc{JAX} optimisations set for \afs{} to \texttt{FastFourierSAT} to provide a more level playing field, and note that the parity learning results shown for \texttt{FastFourierSAT} are better than those presented in their paper due to bugs in the published code we have corrected.
Key observations:
\begin{itemize}
  \item \afs{} achieves equivalent performance in the worst-case, and otherwise universally improves upon \texttt{FastFourierSAT}, demonstrating that our optimisations strictly improved upon the proof-of-concept on comparable benchmarks.
  \item \afs{} exhibits an obviously lower baseline times than \texttt{FastFourierSAT}, indicating more efficient pre-processing, GPU kernel optimisation/compilation, and general overheads. For fairness, we take the best time of several runs for each to allow for disk latency and compilation caches to be populated.
  \item \afs{} consumes substantially less GPU memory by storing only minimal auxiliary data (\eg highly optimised DFT matrices, clause and literal arrays, \etc) and computing closures over this data, enabling the compilers to better optimise which subsequently enables larger batch sizes.
\end{itemize}

\subsection{GPU Memory and Throughput Characteristics}
\label{sec:gpumem}
\begin{figure}[!htbp]
  \centering
  \begin{subfigure}{\textwidth}
    \centering
    \includegraphics[width=0.8\linewidth]{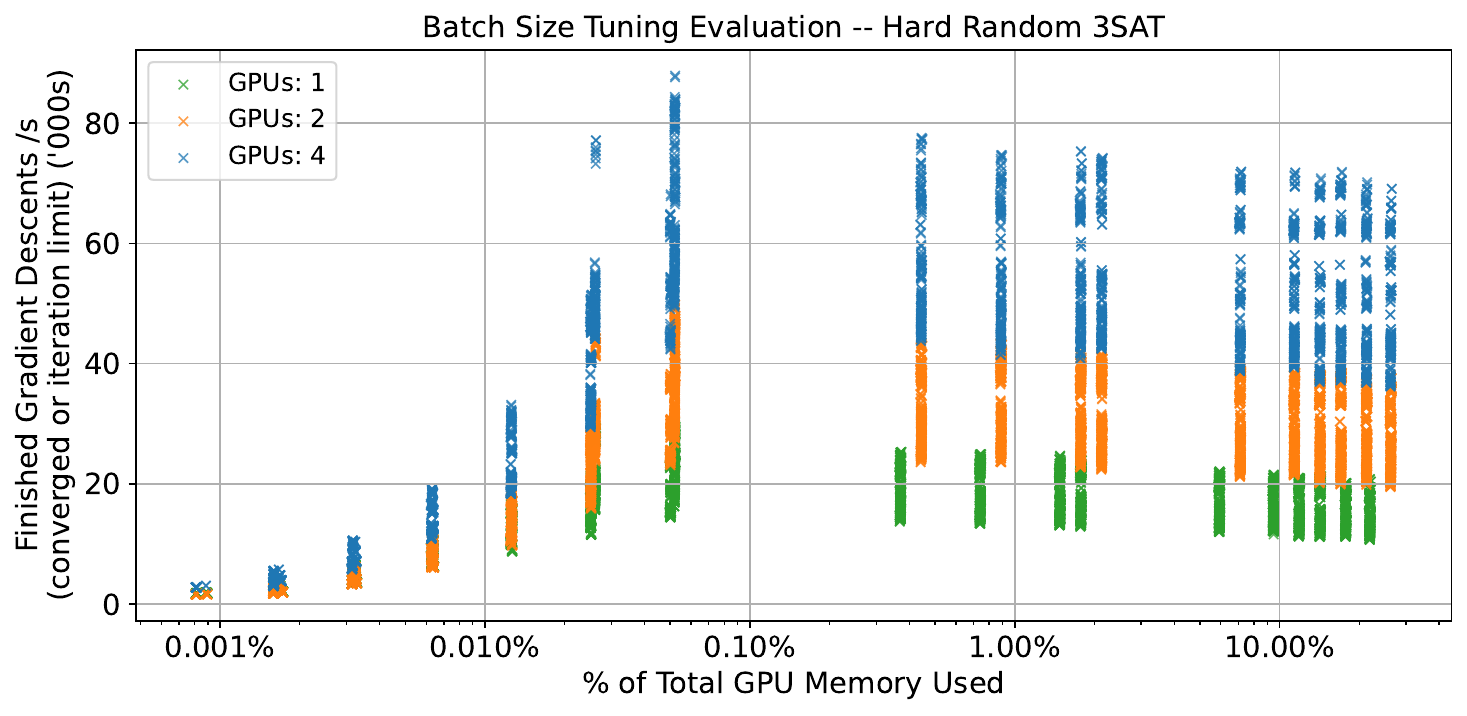}
  \end{subfigure}
  \begin{subfigure}{\textwidth}
    \begin{center}
    \includegraphics[width=0.8\linewidth]{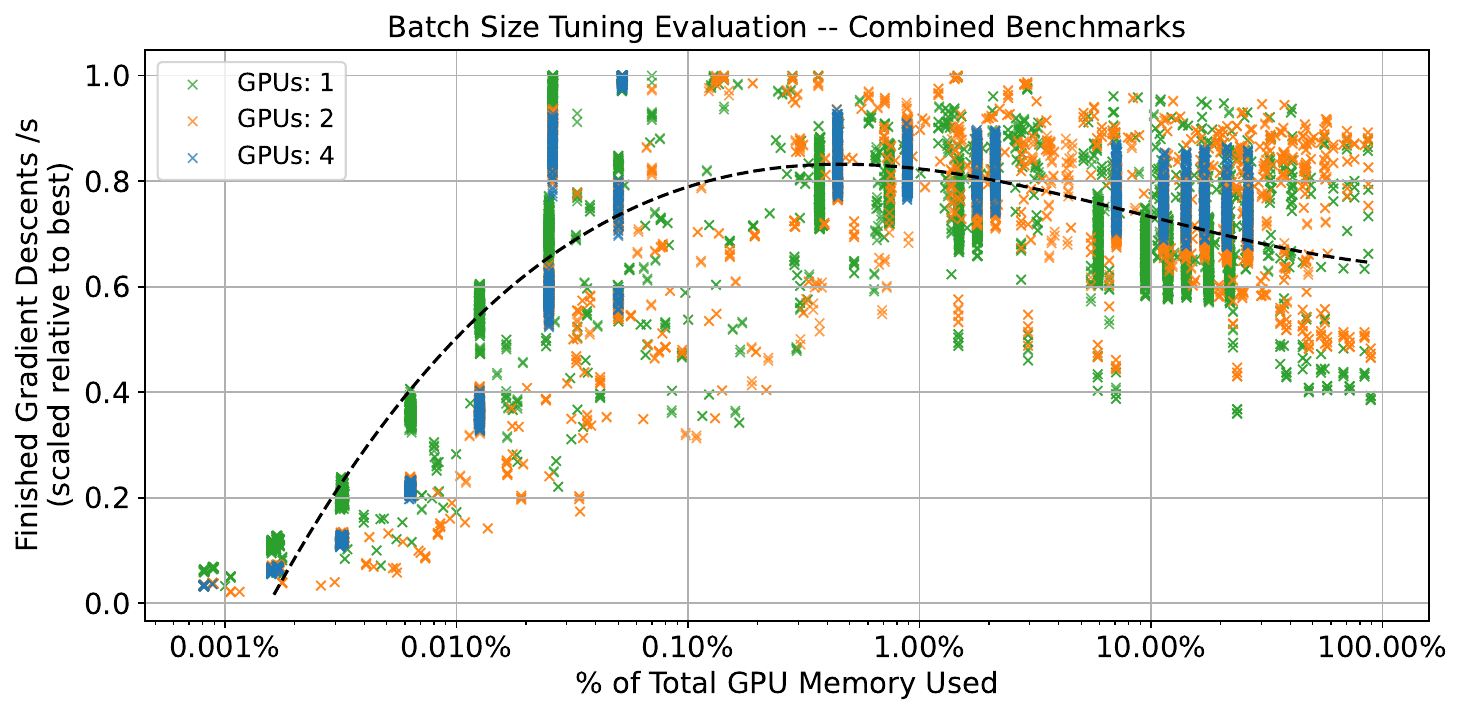}
    \end{center}
  \end{subfigure}
  \caption{Throughput vs measured memory consumption across various problem classes.\\
  \emph{(Subfig. a)} shows near linear scaling when parallelising across several accelerators on various hard random 3SAT problems, with the occasional exception due to properties of individual test cases.\\
  \emph{(Subfig. b)} shows scaled throughput for various benchmarking problems, confirming the dynamics of throughput vs. batch size (memory consumption). A polynomial trendline peaks between $0.1\%$ and $1\%$.}
  \label{fig:mem_all}
\end{figure}

During testing it was noticed that as total GPU memory utilisation increased (reflecting larger batch sizes), the total number of completed gradient descents decreased both relatively (with respect to the amount of memory consumed) and in some cases absolutely. We investigate the total number of completed searches we can take per unit time, a measure we call \emph{throughput}, as a function of GPU memory consumption and GPUs available across several problem domains (Figure~\ref{fig:mem_all}). We observe clearly that throughput peaks when the memory used is approximately $0.1\%$--$1\%$ of total GPU memory consumed. When exploring this effect across various GPUs, we find this percentage varies, but upon closer inspection the value is more tightly correlated twice the cache memory available on the device. We observe a logarithmic decay (Figure~\ref{fig:mem_all}b) in throughput beyond this peak, rather than the monotonic increases one might expect when parallelising a problem up to memory saturation.

Across various domains we also observe the optimal batch size (and subsequently peak throughput) is governed primarily by maximum constraint length rather than the absolute number of variables or constraints. This is dominated ultimately by the requirement to compute $O(k^2)$ DFT matrices for $k$ length clauses. As the size of the DFT grows quadratically it rapidly cuts the amount of memory usable for the remainder of the algorithm, resulting in lower batch sizes. We hypothesise that the throughput peak corresponds to saturation of low-level processing core caches rather than main memory, whereby optimal pipelining is achieved with next to no cache invalidation or page-faulting. Supporting this, the ratio of total cache to total memory on the V100 architecture is on the order of $0.1\%$---consistent with the observed peak locations.

This finding has practical implications: for problems with long constraints, smaller batch sizes yield higher throughput than larger ones that exceed cache capacity. For deployment, \afs{} provides batch-size tuning that balances throughput against total search coverage. As indicated in \S\ref{subsec:fp}, long constraints also degrade numerical stability. The combination of both of these effects suggests that future improvements will look to trade-off the compactness of single constraints for the efficiency gains of constraint decompositions.

\textsc{JAX}'s sharding mechanism distributes batched valuations across available GPUs in a compute-follows-data paradigm. Since gradient descents are independent across each assignment in the batch, communication overhead is minimal. We observe near-linear scaling in throughput with increasing GPU count across all tested problem domains and Figure~\ref{fig:mem_all}a demonstrates this for hard random 3SAT problems. The absence of an impacting overhead cost confirms the suitability of CLS for multi-accelerator deployment.

%----------------------------------------------------------------------------------------
%  LIMITATIONS & CONCLUSION
%----------------------------------------------------------------------------------------

\section{Conclusion and Future Work}

\afs{} advances the state of CLS-based SAT solving from proof-of-concept to a practical, extensible tool. By engineering support for heterogeneous pseudo-Boolean constraints, achieving substantial performance improvements over the baseline, and demonstrating scalable multi-GPU execution, \afs{} establishes a foundation for GPU-accelerated SAT solving in accelerator-rich environments. The identification of precision-driven constraint-length limits and cache-driven throughput characteristics provides actionable guidance for deployment. Future development will target adaptive constraint weighting, integration with systematic solvers via decomposition frameworks, and exploitation of emerging higher-precision accelerator arithmetic.

\paragraph*{Constraint length ceiling and memory consumption.} As established in \S\ref{subsec:fp} and \S\ref{sec:gpumem}, floating-point precision limits constraint length to $n \approx 50$ on 64-bit hardware and dominates memory consumption. Problems with naturally longer constraints (e.g., global cardinality constraints in large graph colouring) can be addressed by decomposing these into shorter equivalent constraints. For all but EK and CARD, there are simple linear decompositions to $O(k/50)$ equivalent PB constraints of the same type, preserving some of the benefits of native PB representation.

\paragraph*{Incompleteness.} \afs{} is an incomplete solver: it cannot prove unsatisfiability, as doing so would require solving the global optimisation problem to certifiable optimality. However, \afs{} functions naturally as a MaxSAT solver, providing best-effort solutions with unsatisfied-constraint counts. Unbounded global optimisation methods are a potential consideration, and methods involving Moreau envelopes~\cite{cen2025massively} or unconstrained formulations with extreme penalties~\cite{zhang2025thinkingboxhybridsat}.

\paragraph*{Second-order methods.} The multi-linear objective is thoroughly populated with saddle points, at which first-order PGD can stall. Bounded second-order methods (e.g., L-BFGS-B) would help escape saddle points, but existing implementations in the \textsc{JAX} ecosystem---notably \textsc{JAXOpt}~\cite{jaxopt_implicit_diff} and \textsc{Optimistix}~\cite{optimistix2024}---do not correctly respect bounds or are not production-ready. Higher order methods will necessarily decrease throughput due to memory required to compute and store higher order gradients.

%----------------------------------------------------------------------------------------
%  BIBLIOGRAPHY
%----------------------------------------------------------------------------------------

\bibliography{bib}

@misc{zhang2025thinkingboxhybridsat,
      title={Thinking Out of the Box: Hybrid SAT Solving by Unconstrained Continuous Optimization}, 
      author={Zhiwei Zhang and Samy Wu Fung and Anastasios Kyrillidis and Stanley Osher and Moshe Y. Vardi},
      year={2025},
      eprint={2506.00674},
      archivePrefix={arXiv},
      primaryClass={cs.LO},
      url={https://arxiv.org/abs/2506.00674}, 
}

@book{odonnell14, 
    place={Cambridge}, 
    title={Analysis of Boolean Functions}, 
    publisher={Cambridge University Press}, 
    author={O'Donnell, Ryan}, 
    year={2014}
}

@article{cen2025massively,
  title={Massively parallel continuous local search for hybrid SAT solving on GPUs},
  author={Cen, Yunuo and Zhang, Zhiwei and Fong, Xuanyao},
  journal={Proceedings of the AAAI Conference on Artificial Intelligence},
  volume={39},
  number={11},
  pages={11140--11149},
  year={2025}
}

@article{jaxopt_implicit_diff,
  title={Efficient and Modular Implicit Differentiation},
  author={Blondel, Mathieu and Berthet, Quentin and Cuturi, Marco and Frostig, Roy
   and Hoyer, Stephan and Llinares-L{\'o}pez, Felipe and Pedregosa, Fabian
   and Vert, Jean-Philippe},
  journal={arXiv preprint arXiv:2105.15183},
  year={2021}
}

@article{optimistix2024,
    title={Optimistix: modular optimisation in JAX and Equinox},
    author={Jason Rader and Terry Lyons and Patrick Kidger},
    journal={arXiv:2402.09983},
    year={2024},
}

@ARTICLE{2020SciPy-NMeth,
  author  = {Virtanen, Pauli and Gommers, Ralf and Oliphant, Travis E. and Haberland, Matt and Reddy, Tyler and Cournapeau, David and Burovski, Evgeni and Peterson, Pearu and Weckesser, Warren and Bright, Jonathan and {van der Walt}, St{\'e}fan J. and Brett, Matthew and Wilson, Joshua and Millman, K. Jarrod and Mayorov, Nikolay and Nelson, Andrew R. J. and Jones, Eric and Kern, Robert and Larson, Eric and Carey, C J and Polat, {\.I}lhan and Feng, Yu and Moore, Eric W. and {VanderPlas}, Jake and Laxalde, Denis and Perktold, Josef and Cimrman, Robert and Henriksen, Ian and Quintero, E. A. and Harris, Charles R. and Archibald, Anne M. and Ribeiro, Ant{\^o}nio H. and Pedregosa, Fabian and {van Mulbregt}, Paul and {SciPy 1.0 Contributors}},
  title   = {{{SciPy} 1.0: Fundamental Algorithms for Scientific Computing in Python}},
  journal = {Nature Methods},
  year    = {2020},
  volume  = {17},
  pages   = {261--272},
  adsurl  = {https://rdcu.be/b08Wh},
  doi     = {10.1038/s41592-019-0686-2},
}

@misc{deepmind2020jax,
  title = {The {D}eep{M}ind {JAX} {E}cosystem},
  author = {DeepMind and Babuschkin, Igor and Baumli, Kate and Bell, Alison and Bhupatiraju, Surya and Bruce, Jake and Buchlovsky, Peter and Budden, David and Cai, Trevor and Clark, Aidan and Danihelka, Ivo and Dedieu, Antoine and Fantacci, Claudio and Godwin, Jonathan and Jones, Chris and Hemsley, Ross and Hennigan, Tom and Hessel, Matteo and Hou, Shaobo and Kapturowski, Steven and Keck, Thomas and Kemaev, Iurii and King, Michael and Kunesch, Markus and Martens, Lena and Merzic, Hamza and Mikulik, Vladimir and Norman, Tamara and Papamakarios, George and Quan, John and Ring, Roman and Ruiz, Francisco and Sanchez, Alvaro and Sartran, Laurent and Schneider, Rosalia and Sezener, Eren and Spencer, Stephen and Srinivasan, Srivatsan and Stanojevi\'{c}, Milo\v{s} and Stokowiec, Wojciech and Wang, Luyu and Zhou, Guangyao and Viola, Fabio},
  url = {http://github.com/google-deepmind},
  year = {2020},
}

@article{kyrillidis2021solving,
  title={Solving hybrid Boolean constraints in continuous space via multilinear Fourier expansions},
  author={Kyrillidis, Anastasios and Shrivastava, Anshumali and Vardi, Moshe Y and Zhang, Zhiwei},
  journal={Artificial Intelligence},
  volume={299},
  pages={103559},
  year={2021},
  publisher={Elsevier}
}

@InProceedings{10.1007/978-3-031-20862-1_6,
author="Burgess, Mark Alexander
and Gretton, Charles
and Milthorpe, Josh
and Croak, Luke
and Willingham, Thomas
and Tiu, Alwen",
editor="Khanna, Sankalp
and Cao, Jian
and Bai, Quan
and Xu, Guandong",
title="Dagster: Parallel Structured Search with Case Studies",
booktitle="PRICAI 2022: Trends in Artificial Intelligence",
year="2022",
publisher="Springer Nature Switzerland",
address="Cham",
pages="75--89",
isbn="978-3-031-20862-1"
}

%----------------------------------------------------------------------------------------
%  APPENDIX
%----------------------------------------------------------------------------------------
\clearpage
\appendix

\section{Input Format Specification}
\label{sec:input_format}

\afs{} accepts problem instances in standard DIMACS CNF or an extended DIMACS-like hybrid format supporting heterogeneous symmetric pseudo-Boolean constraints.

\paragraph*{DIMACS-like PB constraint grammar}

The hybrid PB problem grammar is defined as follows, where tokens are whitespace-separated:

\begin{bashblock}
<file>      ::= <line>*
<line>      ::= <comment> | <problem> | <constraint>
<comment>   ::= ('c' | '*') <text>
<problem>   ::= 'p' 'cnf' <posint> <posint>
<constraint>::= [<prefix>] <body> '0'
<prefix>    ::= 'h'
<body>      ::= <cnf> | <typed> | <counted>
<cnf>       ::= <lit>+
<typed>     ::= <type_id> <lit>+
<type_id>   ::= 'x' | 'xor' | 'n' | 'nae'
              | 'a' | 'amo' | 'e' | 'eo'
<counted>   ::= <count_id> <lit>+
<count_id>  ::= <exactly> | <card>
<exactly>   ::= ('k' | 'ek') <int>
<card>      ::= ('d' | 'card') <ineq_op> <posint>
<ineq_op>   ::= '<' | '<=' | '>' | '>='
<lit>       ::= <int> \ {0}
<int>       ::= '-'? [1-9] [0-9]*
<posint>    ::= [1-9] [0-9]*
\end{bashblock}

\noindent The type identifiers map to constraint semantics as follows:

\begin{table}[htb]
  \centering
  \begin{tabular}{|Sl|Sl|Sl|}
    \hline
    \textbf{Identifiers} & \textbf{Type} & \textbf{Semantics} \\
    \hline
    (none) & CNF & Standard disjunction \\
    \texttt{x}, \texttt{xor} & \texttt{XOR} & Odd number of literals true \\
    \texttt{n}, \texttt{nae} & \texttt{NAE} & Not all literals equal \\
    \texttt{a}, \texttt{amo} & \texttt{AMO} & At most one literal true \\
    \texttt{e}, \texttt{eo} & \texttt{EO} & Exactly one literal true \\
    \texttt{k}, \texttt{ek} & \texttt{EK} & Exactly $k$ literals true \\
    \texttt{d}, \texttt{card} & \texttt{CARD} & Cardinality threshold \\
    \hline
  \end{tabular}
  \caption{Constraint type identifiers and semantics.}
  \label{tab:input_types}
\end{table}

\noindent For \texttt{CARD} constraints with a plain integer threshold (no operator): positive $k$ defaults to $\geq k$; negative $k$ defaults to $< |k|$. All \texttt{CARD} constraints are internally normalised to $\geq$ or $<$ forms.

\clearpage
\paragraph*{Examples}
\begin{bashblock}
c --- Standard DIMACS CNF ---
p cnf 4 2
1 -2 3 0
-1 4 0

c --- Hybrid format (mixed constraints) ---
p cnf 6 7
h 1 -2 3 0           c CNF: (x1 OR NOT x2 OR x3)
h x 1 2 3 0          c XOR: odd parity over {x1,x2,x3}
h nae 1 2 3 0        c NAE: not all of {x1,x2,x3} equal (at least one true & false)
h amo 1 2 3 0        c AMO: at most one true
h eo 4 5 6 0         c EO:  exactly one true
h ek 2 1 2 3 0       c EK:  exactly 2 of {x1,x2,x3} true
h d >=2 1 2 3 0      c CARD: at least 2 true

c --- CARD inequality variants ---
card 2 1 2 3 0       c >= 2 true  (positive k, default >=)
card -3 1 2 3 4 0    c <  3 true  (negative k, default <)
card >2 1 2 3 4 0    c >  2 true
card <=2 1 2 3 0     c <= 2 true
\end{bashblock}

\subsection{Automatic Simplification}

The parser applies the following reductions:
\begin{itemize}
  \item Unit \texttt{EO} and \texttt{CNF} constraints are extracted as unit-propagated prefix assignments.
  \item Unit \texttt{AMO} constraints are discarded (trivially satisfied).
  \item \texttt{CARD}-$1$ constraints are reduced to \texttt{CNF}; \texttt{EK}-$1$ to \texttt{EO}.
  \item \texttt{CARD}-$n$ and EK-$n$ (where $k$ equals the number of literals) are converted to unit prefix assignments.
  \item \texttt{CARD}-$0$ constraints are discarded (trivially satisfied); \texttt{EK}-$0$ constraints yield negated prefix assignments.
  \item Conflicting unit literals, if detected, immediately raise an unsatisfiability error.
\end{itemize}

\end{document}